\documentclass[10pt,twocolumn,letterpaper]{article}

\usepackage[pagenumbers,applications]{wacv}

\usepackage{xcolor}
\definecolor{wacvblue}{rgb}{0.21,0.49,0.74}

\usepackage{xcolor}
\definecolor{wacvblue}{rgb}{0.21,0.49,0.74}
\usepackage[pagebackref,breaklinks,colorlinks,allcolors=wacvblue]{hyperref}
\usepackage{booktabs}
\usepackage{adjustbox}
\usepackage{amsmath,amssymb,amsfonts}
\usepackage{graphicx}
\usepackage{textcomp}

\def\confName{WACV}
\def\confYear{2027}

\title{PulseGuard: rPPG-Derived Temporal Signals as a Forensic Modality for Talking-Face Deepfake Detection}

\author{Othmane Harraq\\ Temple University\\ {\tt\small othmane.harraq@temple.edu} \and Tamer Aldwairi\\ Temple University\\ {\tt\small aldwairi@temple.edu}}

\begin{document}
\maketitle

\begin{abstract}
Talking-face (TF) deepfake generation synthesizes photorealistic facial video from a static source image and an audio signal, producing forgeries that current image-based detectors consistently fail to identify. Unlike face-swap manipulation, TF synthesis has no underlying real video from which to inherit physiological characteristics, motivating remote photoplethysmography (rPPG) as a detection modality for this forgery category. We propose PulseGuard, a detection framework that extracts per-window rPPG waveforms via RhythmFormer and trains a suite of lightweight classifiers to distinguish real from synthesized traces. Evaluated on the TF subset of Celeb-DF++ under a strict subject-independent protocol, where test identities are completely separated from training identities, our 1D ResNet achieves a window-level AUC of 0.826 at EER 25.1\% (0.842 pooled to video level, the granularity at which DeepFakesON-Phys is scored) while operating exclusively on the extracted temporal signal, and retains AUC 0.754 on generators withheld entirely from training. We document a controlled reproduction study of DeepFakesON-Phys, the representative prior rPPG detector, demonstrating degradation from AUC 0.999 on legacy face-swap data to 0.576 on the TF subset of Celeb-DF++. We further show that detection difficulty is strongly method-dependent: AUC ranges from 0.937 (Real3DPortrait) to 0.617 (IP-LAP) across the seven TF generators, with the ordering broadly preserved under per-method training. This spread is stable across training regimes rather than reflecting evaluation noise, and constitutes the primary empirical contribution of the work. We report these as empirical baselines and do not establish that the discriminative content is cardiac in origin.
\end{abstract}

\section{Introduction}

Deepfake video generation has advanced to the point where forgeries routinely deceive both human viewers and automated forensic systems. The Celeb-DF++ benchmark \cite{li2025celebdf} catalogues 22 contemporary synthesis methods organized into three categories: face-swap (FS), face-reenactment (FR), and talking-face (TF). Among these, TF presents a distinctive forensic challenge. TF methods, including SadTalker \cite{zhang2023sadtalker}, IP-LAP \cite{zhong2023iplap}, AniTalker \cite{liu2024anitalker}, EDTalk \cite{tan2024edtalk}, Real3D-Portrait \cite{ye2024real3d}, EchoMimic~\cite{chen2025echomimic}, and FLOAT~\cite{ki2025float}, synthesize the entire facial appearance from a single static source frame conditioned on audio, leaving no real video substrate from which physiological characteristics can be inherited.

This generative property has direct consequences for forensic strategy. In FS forgeries, the source video's spatiotemporal structure, including subtle skin-surface reflectance variations induced by pulsatile blood flow, is partially preserved beneath the swapped region. Remote photoplethysmography (rPPG) methods, which recover cardiac pulse waveforms from these microscopic color variations \cite{verkruysse2008rppg, wang2017rppg}, are therefore working against a residual real physiological signal in FS settings. In TF synthesis, no such substrate exists. Any apparent physiological signal in a TF-generated video must be produced de novo by the generator itself, and we are not aware of any TF generator that explicitly models cardiac signal.

Prior rPPG-based deepfake detectors, FakeCatcher \cite{ciftci2020fakecatcher}, DeepFakesON-Phys \cite{hernandez2021deepfakeson}, DeepRhythm \cite{qi2020deeprhythm}, and related rhythmic-pattern work \cite{liang2021identifying}, were developed and evaluated exclusively on FS data, where rPPG is theoretically least advantaged. The TF detection literature, meanwhile, is dominated by audio-visual approaches that exploit lip-audio temporal inconsistency \cite{liu2024lips, haliassos2021lips}. To our knowledge, no prior work applies rPPG-based detection specifically to TF forgeries, where the physiological argument is strongest.

We address this gap with the following contributions:
\begin{itemize}
\item PulseGuard, to our knowledge the first systematic evaluation of rPPG-derived signals for TF deepfake detection, leveraging RhythmFormer \cite{zou2025rhythmformer} for waveform extraction and lightweight 1D classifiers for discrimination, achieving AUC (Area Under the ROC Curve) 0.826 using only the extracted temporal signal.
\item Controlled reproduction of DeepFakesON-Phys on Celeb-DF++ TF data, documenting an AUC degradation from 0.999 (original FS evaluation) to 0.576, quantifying the distribution shift between 2020-era FS training data and contemporary TF forgeries.
\item Method-stratified analysis across all seven TF generators revealing a stable AUC range of 0.617--0.937 in which the three most detectable generators synthesize pixels from learned latent representations and the two least detectable condition on facial landmarks, with FLOAT the exception, providing an empirical characterization of the TF generation landscape.
\item Subject-independent evaluation protocol with disjoint celebrity identity partitions, producing a baseline that avoids cross-identity leakage.
\item Leave-one-generator-out evaluation, showing that every generator remains detectable above chance when withheld from training (mean AUC 0.754), while transfer loss varies from 0.020 to 0.206 across generators.
\end{itemize}
Our code and results are publicly available at \url{https://github.com/AI-Advanced-Vision-Forensics-Lab/rPPG-TalkingFaceDetector}; the underlying video is distributed by the Celeb-DF++ authors.

\section{Related Work}

\subsection{Talking-Face Synthesis}\label{sec:tfsynth}

TF synthesis methods animate a still source image with audio to produce a target-speaker video, spanning strategies from learned 3D or disentangled-motion synthesis to direct source-frame warping. SadTalker \cite{zhang2023sadtalker}, EDTalk \cite{tan2024edtalk}, and Real3D-Portrait \cite{ye2024real3d} generate pixels from learned 3D or motion-disentangled representations; AniTalker \cite{liu2024anitalker} disentangles motion from appearance via a universal motion representation; FLOAT~\cite{ki2025float} performs flow matching in a learned motion latent space; and EchoMimic~\cite{chen2025echomimic} and IP-LAP \cite{zhong2023iplap} condition generation on facial landmarks, with IP-LAP additionally conditioning on appearance priors drawn from the source frame. Across all seven methods, the fundamental operational property is shared: the output video has no temporally coherent real video underlying it, precluding any inheritance of source-video physiology.

\subsection{Remote Photoplethysmography}

rPPG recovers cardiac pulse signals from subtle periodic color variations in facial skin caused by pulsatile blood flow \cite{verkruysse2008rppg}. Classical approaches use handcrafted signal decomposition \cite{dehaan2013rppg, wang2017rppg}. Deep learning methods have substantially improved robustness: PhysNet \cite{yu2019physnet} and variants apply 3D convolutions to spatiotemporal face volumes; PhysFormer++ \cite{yu2023physformer} introduces transformer architectures to capture long-range temporal dependencies. RhythmFormer \cite{zou2025rhythmformer} proposes a periodic sparse attention mechanism that learns to focus on physiologically relevant temporal positions, achieving state-of-the-art performance across PURE, UBFC-rPPG, COHFACE, VIPL-HR and MMPD.

\subsection{rPPG-Based Deepfake Detection}

FakeCatcher \cite{ciftci2020fakecatcher} pioneered the use of rPPG for deepfake forensics, classifying videos via statistical features derived from spatially pooled rPPG signals. DeepFakesON-Phys \cite{hernandez2021deepfakeson} extended this direction with a convolutional attention network operating on facial regions, reporting AUC 0.999 on Celeb-DF-v2. DeepRhythm \cite{qi2020deeprhythm} proposed attentional heartbeat rhythm analysis via dual-spatial-temporal attention maps. Liang and Deng~\cite{liang2021identifying} demonstrated that rPPG rhythmic patterns are discriminative not only for detection but also for forgery categorization. Beyond rPPG, other temporal cues have been applied to face-swap detection, including optical flow \cite{amerini2019opticalflow} and temporal dropout in 3D CNNs \cite{zhang2021detecting}. All of the above approaches were developed and evaluated on FS data. None targets TF synthesis, where the absence of a real physiological substrate motivates rPPG-based discrimination. 

D'Amelio et al.~\cite{damelio2023cautionary} examine the reliability of rPPG for deepfake detection on FS data under varying compression, and report that detection accuracy does not track rPPG signal quality as expected. Their analysis is confined to FS forgeries, where a residual real physiological signal is present; the TF setting we address has no such substrate.

\subsection{Talking-Face Deepfake Detection}

Dedicated TF detection methods have been proposed primarily from an audio-visual perspective. Haliassos et al.~\cite{haliassos2021lips} demonstrated that lip motion features learned for speech recognition are transferable to forgery detection. Liu et al.~\cite{liu2024lips} explicitly model the temporal inconsistency between audio and visual lip motion in lip-syncing deepfakes, achieving strong results on several benchmarks. Datta et al.~\cite{datta2025vtt} exploit spatial-temporal patterns in the mouth region via a vision temporal transformer. These approaches uniformly require synchronized audio. In contrast, our method operates on the RGB visual signal alone, making it complementary to and applicable in audio-absent or audio-compromised settings.

\subsection{The Celeb-DF++ Benchmark}

Celeb-DF++ \cite{li2025celebdf} spans face-swap (FS), face-reenactment (FR), and talking-face (TF) forgeries, with each category populated by several recent synthesis methods and 22 methods in total. The benchmark's own evaluation finds detector performance uneven across categories: FS forgeries remain the most reliably detected, while FR and TF fall substantially lower, which is what motivates our focus on TF. The best single-model result reported in the benchmark~\cite{li2025celebdf} is 83.0\% AUC under GF-eval, obtained by Effort~\cite{yan2025orthogonal} trained on Celeb-DF-v2 and evaluated across all three forgery categories.

\section{Methodology}

\subsection{Problem Formulation}

We cast TF deepfake detection as a binary classification problem over per-window extracted waveforms. Formally, let $W = \{w_1, \ldots, w_N\}$ denote a corpus of 160-frame facial video windows, each assigned a binary label $y_i \in \{0, 1\}$ inherited from its source video and indicating real ($y=0$) or TF-synthesized ($y=1$). Our goal is to learn a classifier $f: \mathbb{R}^{160} \to [0,1]$ that maps a per-window extracted waveform $\mathbf{x} \in \mathbb{R}^{160}$ to a fake probability. Unless noted otherwise, all reported metrics are computed at the window level.

\begin{figure}[t]
\centering
\adjustbox{trim=0.2cm 4cm 0.2cm 3.5cm, clip, width=0.475\textwidth}{%
  \includegraphics{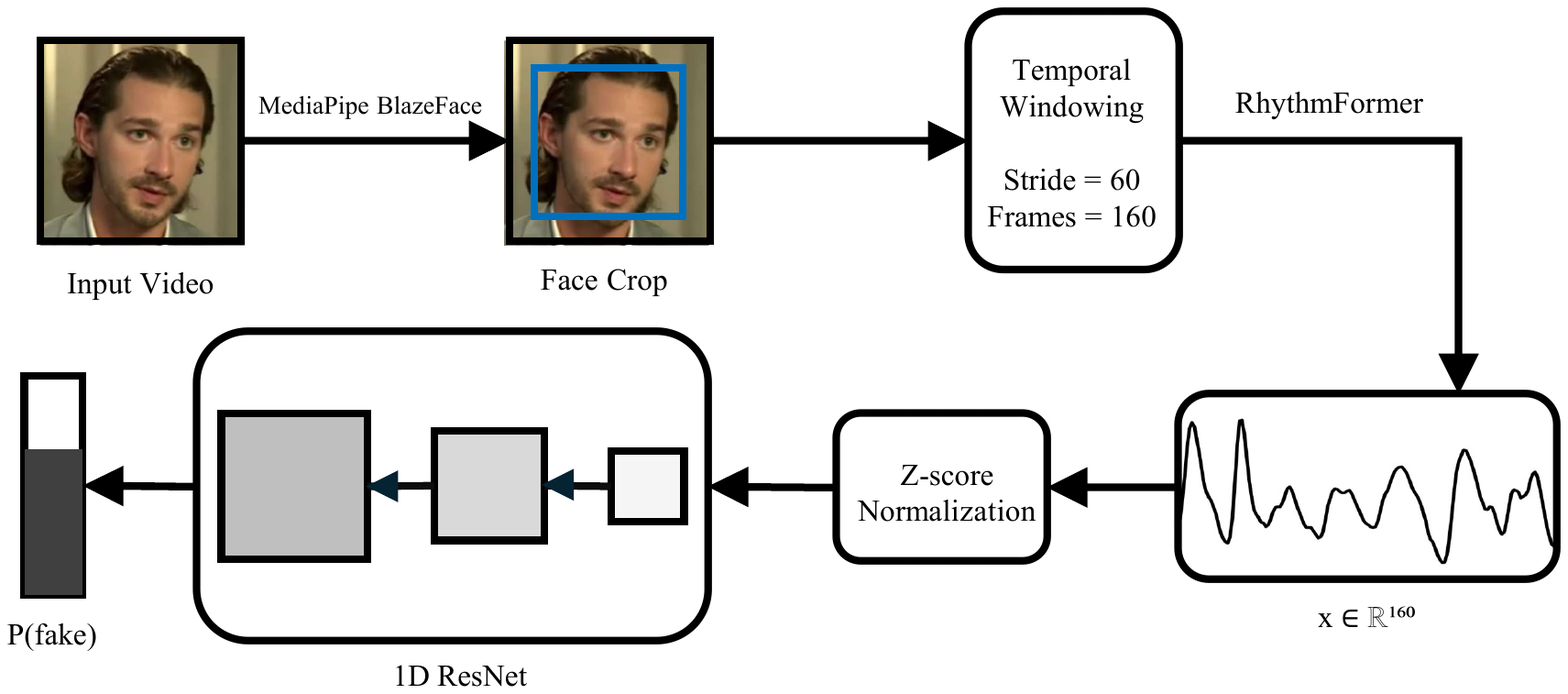}%
}
\caption{Overview of the proposed rPPG-based talking-face detection pipeline. Input video frames are processed by MediaPipe BlazeFace for face detection and cropped with a 1.5$\times$ bounding box expansion. Temporal windowing produces 160-frame clips at stride $s=60$. RhythmFormer extracts a pulse waveform $\mathbf{x} \in \mathbb{R}^{160}$, which is per-window z-score normalized before classification by the 1D ResNet (240K parameters).}
\label{fig:pipeline}
\end{figure}

\subsection{Waveform Extraction}

\textbf{Face Detection.} We replace the RhythmFormer reference pipeline's RetinaFace detector with the MediaPipe BlazeFace short-range model~\cite{bazarevsky2019blazeface} (Tasks API), improving robustness across the head-pose variation common in Celeb-DF++. Face bounding boxes are expanded by a factor of 1.5 with center preservation. The full pipeline is illustrated in Fig.~\ref{fig:pipeline}.

\textbf{Temporal Windowing.} For each video, frames are resampled to yield a clip of exactly 160 frames. For videos of native length $T < 160$ frames (minimum $T \geq 60$), frames are selected via linear interpolation of frame indices, preserving approximate temporal phase relationships. For a video $V$ of length $T \geq 160$ frames, we extract a set of windows $\{W_k\}$ of fixed length $L = 160$ using a sliding stride $s$. The starting frame index of the $k$-th window is given by
\begin{equation}
\label{eq:windowing}
    \text{start}(k) = k \cdot s, \quad k = 0, 1, \ldots, \left\lfloor \frac{T - L}{s} \right\rfloor,
\end{equation}
such that window $W_k$ consists of frames $V[\text{start}(k) : \text{start}(k) + L]$. Section~\ref{sec:temporal-cropping} describes the stride used for real-video windowing.

\textbf{RhythmFormer Inference.} Each 160-frame clip is resized to $128 \times 128$, globally z-score normalized across the spatial-temporal dimensions, and passed through RhythmFormer \cite{zou2025rhythmformer} loaded with the UBFC\_cross checkpoint. RhythmFormer outputs a pulse waveform $\mathbf{x} \in \mathbb{R}^{160}$.

\textbf{Per-Window Normalization.} To remove identity-specific baseline amplitude (a DC component that encodes subject identity), each waveform is independently z-score normalized:
\begin{equation}
\label{eq:zscore}
    \tilde{\mathbf{x}} = \frac{\mathbf{x} - \mu_{\mathbf{x}}}{\sigma_{\mathbf{x}} + \epsilon}, \quad \epsilon = 10^{-6}.
\end{equation}
This operation preserves the oscillation shape and frequency content of the waveform while erasing absolute amplitude and mean, rendering the representation more identity-agnostic.

\subsection{Classifier Architectures}

We evaluate three core architectures of constrained scale (36K, 240K, 69K parameters), designed to characterize the discriminative information available in the extracted waveforms. Two additional Toeplitz-based representations were tested and are discussed in Section~\ref{sec:ablation-failed} as a negative result.

\textbf{1D Residual Network (1D ResNet).} Three stages of basic residual blocks are used, with configuration $(2,2,2)$ and channel widths 32, 64, 128. Each basic block consists of two Conv1d layers with batch normalization and a skip connection. A stem convolution and global average pooling precede the final linear classifier. This architecture totals 240K parameters and serves as the primary classifier throughout our experiments.

\textbf{1D Convolutional Network (1D CNN).} Three sequential Conv1d blocks with channel dimensions $1 \to 32 \to 64 \to 128$ and kernel sizes 7, 5, 3 are applied in sequence. Each block performs bias-free convolution, batch normalization, ReLU activation, and dropout. A global average pooling operation is followed by a linear projection to a scalar logit. This architecture totals 36K parameters.

\textbf{1D Transformer.} A patch size of 8 yields 20 tokens per sequence, with a model dimension of 64, 4 attention heads, and 2 encoder layers. The architecture uses learned positional embeddings, a learnable [CLS] token, and pre-layer normalization. This architecture totals 69K parameters.

\subsection{Training Protocol}

All classifiers are trained with a weighted binary cross-entropy loss to address class imbalance. For a batch of $N$ samples with ground-truth labels $y_i$ and predicted probabilities $\hat{y}_i$, the loss is defined as
\begin{equation}
\label{eq:wbce}
    \mathcal{L}_{\text{WBCE}} = -\frac{1}{N} \sum_{i=1}^{N} \Big[ w_+ \cdot y_i \log(\hat{y}_i) + (1-y_i) \log(1-\hat{y}_i) \Big],
\end{equation}
where the positive class weight $w_+ = n_- / n_+$ is the ratio of negative (real) to positive (fake) samples in the training set, compensating for the class imbalance in our corpus.

We use AdamW~\cite{loshchilov2019adamw} with weight decay for 30 epochs at batch size 64. The best-performing configuration, confirmed via cross-validation, uses lr $= 10^{-3}$, wd $= 5 \times 10^{-4}$, and dropout $= 0.5$ for the 1D ResNet; lr $= 5 \times 10^{-4}$, wd $= 10^{-3}$, and dropout $= 0.3$ for the 1D CNN; and lr $= 10^{-3}$, wd $= 10^{-4}$, and dropout $= 0.3$ for the Transformer.

Portions of the implementation, including data preprocessing and model training code, were developed with the assistance of an AI coding tool; all code was reviewed and validated by the authors.

\subsection{Ablation Study and Failed Representations}

\label{sec:ablation-failed}
We additionally evaluated two representations in which the waveform is expanded into a symmetric Toeplitz matrix 
$\mathbf{T} \in \mathbb{R}^{160 \times 160}$,
\begin{equation}
\label{eq:toeplitz}
    T_{ij} = \tilde{x}_{|i-j|+1}, \quad i,j \in \{1, \ldots, 160\},
\end{equation}
processed either by a three-block 2D convolutional network (32K parameters) or, at patch size 16, by a 90K-parameter Vision Transformer with model dimension 64, 4 attention heads, and 2 encoder layers. Neither improves on the 1D baselines: on the 18-identity eval the 2D CNN reaches AUC $0.772 \pm 0.011$ and the Vision Transformer $0.757 \pm 0.006$, against $0.826 \pm 0.005$ for the 1D ResNet.

We report this as a negative result with an explicit caveat: 
Equation~\eqref{eq:toeplitz} replicates the same 160 values across the matrix without forming any pairwise product, and therefore carries no information beyond the 1D input. A representation that genuinely exposes lag-domain structure requires a nonlinear function of the waveform, such as the autocorrelation
\begin{equation}
\label{eq:autocorr}
    r(\tau) = \sum_{t} \tilde{x}_t \, \tilde{x}_{t+\tau},
\end{equation}
which we did not test. The failure of Equation~\eqref{eq:toeplitz} is thus a property of the construction rather than evidence that lag-domain features are uninformative for this task.

\section{Experimental Setup}

\subsection{Dataset}

We evaluate on the TF subset of Celeb-DF++ \cite{li2025celebdf}, which contains 590 real videos from 59 unique celebrity identities and 20,279 TF-forged videos spanning seven synthesis methods. Eleven real source videos yield no usable waveform, leaving 579 sources. Stride-60 temporal windowing of these 579 videos produces 2,371 real waveforms, whereas each fake video contributes a single waveform.

\subsection{Subject-Independent Data Partition}

To prevent cross-identity physiological leakage, we partition data at the identity level. The 59 unique celebrity identities are assigned to disjoint splits: 41 identities to training, 9 to validation, and 9 to test. All videos from a given identity are assigned exclusively to one split, verified before any experiments are run. The partition is summarized in Table~\ref{tab:dataset_splits}.

\begin{table}[ht]
\caption{Dataset Split Statistics}
\label{tab:dataset_splits}
\centering
\renewcommand{\arraystretch}{1.2}
\begin{tabular}{lcccc}
\toprule
\textbf{Split} & \textbf{Identities} & \textbf{Real} & \textbf{Fake} & \textbf{Total} \\
\midrule
Train & 41 & 1,675 & 14,120 & 15,795 \\
Val   & 9  & 300   & 2,693  & 2,993  \\
Test  & 9  & 396   & 3,466  & 3,862  \\
\bottomrule
\end{tabular}
\end{table}

The test identities (id0, id4, id6, id11, id13, id16, id23, id27, id54) are completely withheld from all training and hyperparameter selection procedures.

\subsection{Temporal Cropping}\label{sec:temporal-cropping}

All real videos undergo temporal windowing at stride $s = 60$ (Equation~\eqref{eq:windowing}) at extraction time, where the identity-level split is applied afterward to the resulting windows. Fake videos yield one waveform each: those with at least 160 native frames contribute their first 160 frames, and shorter videos are linspace-stretched to 160. Real videos are therefore sampled across their full duration while forgeries are not, an asymmetry we do not control for. This produces 1,675 real training waveforms against 14,120 fake, a ratio of approximately 1:8, compensated by $w_+ \approx 0.119$ in
Equation~\eqref{eq:wbce}. The weight is recomputed from the realized label counts of each training split rather than fixed in advance.

\subsection{Cross-Validation Protocol}

Hyperparameter selection uses 5-fold StratifiedGroupKFold with celebrity identity as the grouping key, ensuring that all temporal-crop waveforms derived from the same source video remain in the same fold and that validation folds contain identities completely absent from training folds.

\subsection{Evaluation Metrics}

We report AUC (area under the ROC curve) as the primary discriminability metric. EER (equal error rate) reports the operating point at which false accept and false reject rates are equal, a standard summary in biometric evaluation. For multi-seed experiments, we report mean $\pm$ standard deviation across 5 random seeds to characterize result stability.

\section{Experiments and Results}

\subsection{Reproduction Study: DeepFakesON-Phys}
To establish a reference baseline and quantify distribution shift between 2020-era face-swap data and contemporary TF forgeries, we execute the official DeepFakesON-Phys pipeline \cite{hernandez2021deepfakeson} on our TF corpus using released model weights and preprocessing code. Several compatibility patches are required for current software dependencies (NumPy 2.x, TensorFlow legacy Keras mode, headless OpenCV). With the official preprocessing unchanged, we measure a video-level AUC of 0.576 on our TF corpus, compared to 0.999 reported by the original authors on Celeb-DF-v2 face-swap data. Performance is additionally unstable across generators, ranging from 0.716 (FLOAT) to 0.384 (SadTalker); on three of the seven methods the detector performs at or below chance. The pipeline requires raw video and scores a strict subset of our test forgeries, 3,171 of 3,466, and 100 real videos against our 99. Its per-generator ordering does not correlate with ours: FLOAT is its best case and our fifth, SadTalker its worst and our third. This is consistent with the two pipelines keying on different artifacts, and is a reason to treat the ordering of Section~\ref{sec:permethod} as specific to this extraction pipeline.

The 0.423-point degradation reflects two compounding factors. First, there is a severe distribution shift: DeepFakesON-Phys was trained exclusively on face-swap forgeries, which partially preserve source-video physiology, whereas TF synthesis produces different waveform statistics. Second, the Haar cascade face detector in the reference pipeline exhibits limited robustness on the non-frontal head poses common in TF outputs, producing degraded crops that corrupt physiological signal estimation.

\subsection{Technique Isolation}

Table~\ref{tab:technique} isolates two preprocessing choices under 5-fold StratifiedGroupKFold on the training identities. Per-window z-score normalization adds $+0.005$ AUC and is the only condition that improves on the baseline in all five folds. Waveform augmentation, here adding Gaussian noise at $0.1\times$ the per-window standard deviation with amplitude scaling $\mathcal{U}(0.8, 1.2)$, contributes nothing measurable in isolation ($+0.0002$) and nothing on top of z-score normalization ($-0.005$ against z-score alone, within one cross-fold standard deviation and with roughly twice the cross-fold standard deviation). All other experiments in this paper therefore use z-score normalization without augmentation.

\begin{table}[t]
\caption{Technique isolation. 5-fold StratifiedGroupKFold on the training identities, 1D ResNet.}
\label{tab:technique}
\centering
\begin{tabular}{lcc}
\toprule
\textbf{Condition} & \textbf{CV AUC} & \textbf{EER} \\
\midrule
Baseline (neither)  & $0.786 \pm 0.005$ & 28.7\% \\
Z-score only        & $0.791 \pm 0.006$ & 28.2\% \\
Augmentation only   & $0.786 \pm 0.007$ & 29.0\% \\
Both                & $0.786 \pm 0.010$ & 28.8\% \\
\bottomrule
\end{tabular}
\end{table}

\subsection{Main Results}
Table~\ref{tab:main_results} presents the primary evaluation on the 18-identity eval set (validation and test identities combined), averaged across 5 random seeds. The 1D ResNet yields window-level AUC $0.826 \pm 0.005$ at EER 25.1\%, exceeding the 1D CNN by 0.016 and the Transformer by 0.020. Pooling window scores by mean over each source video, the granularity at which DeepFakesON-Phys is scored, gives $0.842 \pm 0.006$ at EER 23.4\%. The 1D CNN reaches $0.846 \pm 0.009$ at this granularity, a difference within seed variance, so the two should be read as tied rather than ranked; we retain the 1D ResNet as the primary architecture, as it leads at window level and is the model used throughout the per-method analyses.

The spread across architectures is 0.020 AUC despite a seven-fold range in parameter count, and is small relative to the 0.320 spread across generation methods (Section~\ref{sec:permethod}). Classifier capacity is not the binding constraint on this task; the quality of the extracted signal is. We are not aware of a contemporary TF detector evaluated under an identity-disjoint protocol on this corpus, so no same-protocol comparison to a current method is available.

\begin{table}[ht]
\caption{Main Results --- 18-Identity Eval (5 Seeds). Video AUC pools window scores by mean per source video. DeepFakesON-Phys is scored on the 9-identity test split only.}
\label{tab:main_results}
\centering
\small
\renewcommand{\arraystretch}{1.2}
\begin{tabular}{lcc}
\toprule
\textbf{Method} & \textbf{Window} & \textbf{Video} \\
\midrule
DeepFakesON-Phys~\cite{hernandez2021deepfakeson} & --- & 0.576 \\
\midrule
\multicolumn{3}{l}{\textit{PulseGuard}} \\
\quad 1D ResNet    & $0.826 \pm 0.005$ & $0.842 \pm 0.006$ \\
\quad 1D CNN       & $0.810 \pm 0.002$ & $0.846 \pm 0.009$ \\
\quad Transformer  & $0.806 \pm 0.004$ & $0.815 \pm 0.012$ \\
\bottomrule
\end{tabular}
\end{table}

\subsection{Per-Method Analysis}\label{sec:permethod}
\begin{figure}[t]
\centering
\includegraphics[width=\columnwidth]{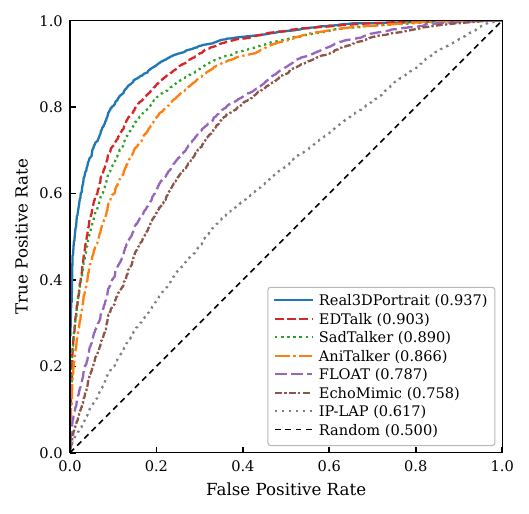}
\caption{ROC curves for the 1D ResNet evaluated per talking-face generation method under the 18-identity subject-independent protocol (5-seed mean). Line style and color together encode method identity for grayscale readability. The AUC spread of 0.320 across methods is stable across training regimes rather than reflecting evaluation noise.}
\label{fig:roc}
\end{figure}

The central empirical finding of this work is the strong dependence of detection difficulty on the specific TF generation method. Fig.~\ref{fig:roc} shows the ROC curves for each method under the 18-identity subject-independent protocol, and Table~\ref{tab:permethod} reports the corresponding AUC values.
\begin{figure}[t]
\centering
\includegraphics[width=\columnwidth]{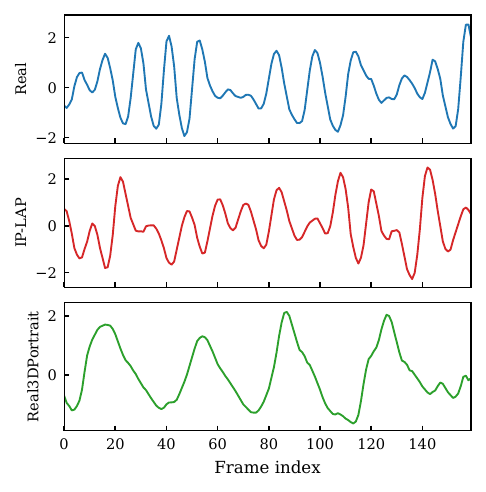}
\caption{Representative traces (z-score normalized) extracted by RhythmFormer from a real video, an IP-LAP fake, and a Real3DPortrait fake. The IP-LAP example shows greater temporal regularity than the Real3DPortrait example. These traces are illustrative and not quantitative evidence.}
\label{fig:waveforms}
\end{figure}
\begin{table}[ht]
\caption{Per-Method Evaluation --- combined-method 1D ResNet, 18-identity protocol (5 seeds)}
\label{tab:permethod}
\centering
\renewcommand{\arraystretch}{1.2}
\begin{tabular}{lcc}
\toprule
\textbf{Method} & \textbf{AUC (mean $\pm$ std)} & \textbf{EER} \\
\midrule
Real3DPortrait & $0.937 \pm 0.009$ & 14.4\% \\
EDTalk         & $0.903 \pm 0.003$ & 18.0\% \\
SadTalker      & $0.890 \pm 0.021$ & 19.0\% \\
AniTalker      & $0.866 \pm 0.013$ & 21.6\% \\
FLOAT          & $0.787 \pm 0.005$ & 28.7\% \\
EchoMimic      & $0.758 \pm 0.014$ & 30.4\% \\
IP-LAP         & $0.617 \pm 0.009$ & 41.1\% \\
\midrule
Mean           & $0.823$           & ---     \\
\bottomrule
\end{tabular}
\end{table}

The AUC spread of 0.320 across methods is far larger than the variance attributable to random seed, consistent with method identity being the dominant factor in detection difficulty. The ordering is broadly stable under isolated per-method training (Table~\ref{tab:isolated}): the three easiest generators and IP-LAP as the hardest are preserved. One inversion occurs mid-range, where EchoMimic overtakes FLOAT under isolated training, consistent with EchoMimic showing the largest gain from a dedicated decision boundary ($+0.122$). EDTalk and SadTalker are separated by 0.013, less than SadTalker's seed standard deviation, and should not be read as reliably ordered.

We considered whether the ordering tracks each generator's synthesis strategy. Six of the seven fit: the three most detectable (Real3DPortrait, EDTalk, SadTalker) synthesize pixels from learned latent representations, and the two least detectable (EchoMimic, IP-LAP) condition on facial landmarks. Real3DPortrait and FLOAT are metadata-indistinguishable; the AUC magnitudes and rank correlations are invariant under exchanging their labels, though the architectural attribution below is not. FLOAT is the exception, performing flow matching in a learned motion latent space (Section~\ref{sec:tfsynth}) yet falling among the harder methods. We report this as an observation rather than a mechanism, since we do not manipulate synthesis strategy experimentally.

IP-LAP is consistently the hardest method in our evaluation. The 1D ResNet reaches AUC $0.617$ under the combined model and $0.710$ when trained on IP-LAP fakes alone (Table~\ref{tab:isolated}), clearly discriminative, but $0.141$ below the next-hardest method (EchoMimic) and $0.320$ below Real3DPortrait. Fig.~\ref{fig:waveforms} shows representative traces. We conjecture that IP-LAP's landmark-guided attention preserves the source frame's identity structure more faithfully than the other methods, and thereby preserves enough temporal color coherence for the extracted waveform to remain periodic rather than becoming absent or noisy. We emphasize that this is a proposed mechanism rather than a measured one: we do not isolate the contribution of the landmark-guided pathway, and an alternative explanation consistent with our data is that IP-LAP's outputs simply retain more of the source frame's compression and color statistics.

\subsection{Isolated Per-Method Training}

To further characterize the learnability of each method's extracted waveform in isolation, we conduct an additional experiment in which the model is trained and evaluated exclusively on one method at a time, using 1,675 real and 1,910--2,055 fake training waveforms depending on method, a near-balanced ratio, under the same subject-independent identity split. Results are summarized in Table~\ref{tab:isolated}.

\begin{table}[ht]
\caption{Isolated Per-Method Training, 18-identity protocol (5 seeds). $\Delta$ compares the isolated 1D ResNet against the same method under the combined-method model (Table~\ref{tab:permethod}), computed at full precision.}
\label{tab:isolated}
\centering
\small
\setlength{\tabcolsep}{5pt}
\begin{tabular}{lccc}
\toprule
\textbf{Method} & \textbf{ResNet} & \textbf{Transf.} & \textbf{$\Delta$} \\
\midrule
Real3DPortrait & $0.988 \pm 0.001$ & $0.980 \pm 0.003$ & $+0.051$ \\
EDTalk         & $0.955 \pm 0.004$ & $0.947 \pm 0.001$ & $+0.053$ \\
SadTalker      & $0.947 \pm 0.002$ & $0.941 \pm 0.004$ & $+0.057$ \\
AniTalker      & $0.924 \pm 0.005$ & $0.910 \pm 0.003$ & $+0.057$ \\
EchoMimic      & $0.880 \pm 0.004$ & $0.842 \pm 0.005$ & $+0.122$ \\
FLOAT          & $0.829 \pm 0.004$ & $0.800 \pm 0.012$ & $+0.041$ \\
IP-LAP         & $0.710 \pm 0.005$ & $0.742 \pm 0.004$ & $+0.092$ \\
\midrule
Mean           & $0.890$ & $0.880$ & $+0.068$ \\
\bottomrule
\end{tabular}
\end{table}

Isolated training improves every method, by $+0.041$ (FLOAT) to $+0.122$ (EchoMimic), with a mean gain of $+0.068$. The two largest gains fall on EchoMimic and IP-LAP, two of the three hardest methods under the combined model.

The combined model therefore pays a measurable cost for joint coverage, disproportionately where generators are hardest to separate, while the endpoints are preserved: IP-LAP remains last at 0.710 and Real3DPortrait first at 0.988. 

\subsection{Leave One Generator Out (LOGO)}\label{sec:logo}

\begin{table}[ht]
\caption{Leave-one-generator-out generalization --- 1D ResNet, 18-identity protocol (5 seeds). The held-out generator contributes no training data. $\Delta$ compares against the combined model trained on all seven (Table~\ref{tab:permethod}), computed at full precision.}
\label{tab:logo}
\centering
\small
\begin{tabular}{lccc}
\toprule
\textbf{Held-out} & \textbf{LOGO AUC} & \textbf{EER} & \textbf{$\Delta$} \\
\midrule
SadTalker      & $0.863 \pm 0.013$ & 21.7\% & $-0.026$ \\
EDTalk         & $0.852 \pm 0.011$ & 23.0\% & $-0.051$ \\
AniTalker      & $0.843 \pm 0.008$ & 24.0\% & $-0.024$ \\
FLOAT          & $0.768 \pm 0.010$ & 30.0\% & $-0.020$ \\
Real3DPortrait & $0.730 \pm 0.038$ & 34.5\% & $-0.206$ \\
EchoMimic      & $0.656 \pm 0.013$ & 38.1\% & $-0.102$ \\
IP-LAP         & $0.567 \pm 0.009$ & 45.6\% & $-0.051$ \\
\midrule
Mean           & $0.754$           & ---    & $-0.069$ \\
\bottomrule
\end{tabular}
\end{table}

Table~\ref{tab:logo} reports leave-one-generator-out results. Every generator remains detectable above chance when withheld from training, with a mean LOGO AUC of 0.754, above the 50--70\% band reported for existing detectors on TF forgeries~\cite{li2025celebdf}, though EchoMimic and IP-LAP fall within it and the protocols are not directly comparable.

Transferability is not the same axis as difficulty. Real3DPortrait is the most detectable generator in-domain (0.937) yet suffers the largest transfer loss ($-0.206$) and the highest seed variance, indicating a strong but idiosyncratic signature. FLOAT, by contrast, is among the harder generators in-domain but transfers almost intact ($-0.020$). The combined model therefore captures a mixture of a shared discriminant, which transfers, and generator-specific artifacts, which do not.

\subsection{Frame-Rate Normalization}\label{sec:fps}

Real videos in Celeb-DF++ are encoded at 30\,fps while six of the seven generators output 25\,fps and EchoMimic 24\,fps, and our extraction pipeline consumes native frames without resampling. A fixed 160-frame window therefore spans 5.3\,s of real footage but 6.4\,s of generated footage at 25\,fps and 6.7\,s for EchoMimic, shifting the frequency content of the extracted waveform by class independently of any physiological difference.

We test the effect by resampling every video to a common 30\,fps timeline before extraction, applying identical treatment to both classes, and re-running the combined-model evaluation. Pooled AUC rises from 0.822 to 0.846 under the original training run, and every generator improves, most substantially IP-LAP (0.617 $\to$ 0.704). The direction is consistent with RhythmFormer's UBFC checkpoint having been trained at 30\,fps. A rate control, separating native-30\,fps from resampled-25\,fps windows of the same real videos, reaches AUC 0.631 (0.481 shuffled), indicating that rate is separable in isolation. The normalized run also excludes 2,225 of the 20,279 forgeries, which the native run stretched to length by linear index interpolation, a path applied almost entirely to the fake class. Removing those clips from the evaluation raises AUC by 0.006 with the ordering unchanged ($\rho = 1.00$), so they were marginally harder fakes rather than a source of inflation.

\section{Discussion}

\subsection{rPPG Extraction as a Forensic Modality for Talking-Face Detection}
Using only signals derived from an rPPG extraction pipeline, with no texture features, no frequency analysis, and no audio, our 240K-parameter 1D ResNet achieves AUC 0.826. We report this as an empirical baseline. Whether the discriminative content is cardiac in origin, or reflects broader temporal color coherence introduced by rendering and encoding, is not established by these experiments.

\subsection{The Real Identity Ceiling} 

Although we use 20,279 fake videos spanning seven methods, all real video comes from only 59 unique celebrities. Per-identity AUC varies substantially across the 18 evaluation identities (standard deviation 0.046 at seed 42), an order of magnitude larger than the 0.005 seed standard deviation of the pooled figure, indicating that identity, not seed, dominates the residual variation.

Expanding the real video pool through multi-dataset fusion or collecting additional real data is the most direct path to closing this gap.

\subsection{Method-Dependent Detectability as a Forensic Property} 

The latent-versus-landmark pattern of Section~\ref{sec:permethod}, which six of seven generators fit, is a testable hypothesis rather than a mechanism. Our leave-one-generator-out results add a qualification: detectability and transferability are distinct axes (Section~\ref{sec:logo}), so a single property of the synthesis strategy does not predict both. Confirming either relationship would require manipulating synthesis strategy while holding other factors fixed, which a seven-method sample cannot do. Even as a hypothesis, the spread offers a basis for prioritizing detector development effort.

\subsection{Limitations and Future Work}

We report these results as an empirical baseline and note an important limitation of their interpretation. Control experiments indicate that the discriminative content of the extracted waveform is not specific to the rPPG extractor: alternative extraction paths recover comparable discriminability, and the signal may reflect low-level temporal appearance structure rather than cardiac content. This is consistent with earlier cautions about the reliability of rPPG-based deepfake detection~\cite{damelio2023cautionary}. Characterising that structure and separating it from rendering and compression artifacts is the subject of ongoing work.

Method labels are taken from the directory organisation of the Celeb-DF++ TF release rather than from any property of the video files themselves. We verified that each labelled set is homogeneous in frame rate, resolution, codec, level and encoder tag, and that six of the seven sets are mutually distinguishable on those fields; FLOAT and Real3DPortrait are not, matching on every field we measured. This consistency is necessary but not sufficient: it does not fix which generator name attaches to which set, which rests on the release's own organisation. The magnitude of the per-method spread does not depend on which name attaches to which set, and neither do the rank correlations reported in Sections~\ref{sec:permethod} and~\ref{sec:fps}. Additionally, the real-to-fake training imbalance, compensated by weighted BCE, may not be fully resolved at threshold-dependent metrics such as precision; future work should explore oversampling strategies for the real class. Furthermore, rPPG extraction is sensitive to skin tone and lighting~\cite{nowara2020meta}, and our protocol does not stratify evaluation across demographic groups; future work should evaluate detection rates across the Fitzpatrick scale.

We further acknowledge the dual-use nature of this work: while our findings advance the detection of talking-face deepfakes, the detailed characterization of each generator's detectability profile, particularly the observation that IP-LAP outputs are the hardest to detect, could inform adversarial synthesis strategies designed to evade physiological detectors. We release our findings in the interest of defensive research, consistent with responsible disclosure norms in the security community.

\section{Conclusion}

We have presented PulseGuard, a framework for evaluating rPPG-derived signals on talking-face deepfake synthesis, achieving AUC 0.826 under a rigorous subject-independent protocol on Celeb-DF++ while operating exclusively on rPPG-derived temporal signals. The central finding of this work is that detection difficulty depends strongly on the specific TF generation method, ranging from AUC $0.937$ (Real3DPortrait) to $0.617$ (IP-LAP). The three most detectable generators synthesize pixels from learned latent representations and the two least detectable condition on facial landmarks, though FLOAT is a latent-space method that nonetheless falls among the harder cases. The ordering is broadly preserved under isolated per-method training, with one mid-range inversion, and under leave-one-generator-out evaluation every generator remains detectable above chance (mean AUC 0.754) though the ordering changes, since transferability and in-domain difficulty are distinct axes. Detection difficulty is therefore strongly associated with generator identity and is not explained solely by the training regime.

These findings establish a methodological foundation for rPPG-based TF detection and identify IP-LAP-style identity-preserving synthesis as the primary open challenge for rPPG-based forensics. Future directions include multi-dataset real video fusion to address the 59-identity ceiling, adversarial rPPG injection to harden detectors against physiologically plausible synthesis, and fusion of rPPG features with complementary visual or frequency-domain signals to close the remaining gap to general-purpose detection performance.

{
    \small
    \bibliographystyle{ieeenat_fullname}
    \bibliography{main}
}

\end{document}